\documentclass{article}

\usepackage[main, preprint]{neurips_2026}

\usepackage[utf8]{inputenc} 
\usepackage[T1]{fontenc}    
\usepackage{hyperref}       
\usepackage{url}            
\usepackage{booktabs}       
\usepackage{amsfonts}       
\usepackage{nicefrac}       
\usepackage{microtype}      
\usepackage{xcolor}         
\usepackage{graphicx}
\usepackage{amsmath}

\newif\ifshowchanges
\showchangesfalse
\ifdefined\HideRevisionHighlights\showchangesfalse\fi
\newif\ifcurrentrevision
\newcommand{\revisioncolor}{\ifshowchanges\ifcurrentrevision\color{red}\else\color{blue}\fi\fi}
\DeclareRobustCommand{\rev}[1]{{\revisioncolor#1}}
\DeclareRobustCommand{\finalrev}[1]{{\currentrevisiontrue\revisioncolor#1}}
\newenvironment{revision}{\revisioncolor}{}
\newenvironment{finalrevision}{\currentrevisiontrue\revisioncolor}{}
\hypersetup{hidelinks}

\makeatletter
\@namedef{revisionbib@houthooft2016vime}{}
\@namedef{revisionbib@eysenbach2019diayn}{}
\@namedef{revisionbib@shyam2019active}{}
\@namedef{revisionbib@mazzaglia2022surprise}{}
\@namedef{revisionbib@pathak2017curiosity}{}
\newcommand{\revisionbibcolor}[1]{%
  \ifshowchanges
    \ifcsname revisionbib@#1\endcsname\color{blue}\else\color{black}\fi
  \fi}
\let\review@lbibitem\@lbibitem
\def\@lbibitem[#1]#2{\revisionbibcolor{#2}\review@lbibitem[#1]{#2}}
\let\review@bibitem\@bibitem
\def\@bibitem#1{\revisionbibcolor{#1}\review@bibitem{#1}}
\makeatother

\usepackage{comment} 

\title{Learning actionable information landscapes for multimodal active sensing in hawkmoths}

\author{%
  Abdelrahman Sharafeldin\thanks{Corresponding author} \\
  Machine Learning Center\\
  Georgia Institute of Technology\\
  Atlanta, GA 30332 \\
  \texttt{abdo.sharaf@gatech.edu} \\
  \And
  Yaqing Wang\\
  School of Physics\\
  Georgia Institute of Technology\\
  Atlanta, GA 30332\\
  \texttt{ywang4587@gatech.edu}\\
  \AND
  Simon Sponberg \\
  School of Physics \\
  Georgia Institute of Technology\\
  Atlanta, GA 30332 \\
  \texttt{sponberg@gatech.edu} \\
  \And
  Hannah Choi\thanks{Corresponding author} \\
  School of Mathematics \\
  Georgia Institute of Technology \\
  Atlanta, GA 30332 \\
  \texttt{hannahch@gatech.edu} \\
}

\begin{document}

\maketitle

\begin{abstract}
Animals integrate information across sensory modalities to guide exploration, suggesting that they maintain internal estimates of where actions are expected to reduce uncertainty. However, how such action-conditioned information maps are learned from multimodal sensory experience remains unclear, as previous studies have been largely confined to unimodal sensing frameworks. We introduce a multimodal predictive-coding framework for learning uncertainty-reduction maps from visual and mechanosensory observations in hawkmoth flower interaction tasks. A variational generative model trained on simulated environments recovers modality-specific information landscapes over the action space, capturing how visual patterns, surface geometry, and mechanosensory cues shape the value of exploratory actions. These learned maps are consistent with several features of hawkmoth behavior, including nonzero tracking offsets, probing along visual patterns, and geometry-dependent exploration. Applying the model to behavioral data from flower tracking and nectary search, we find that natural trajectories tend to occupy regions of high uncertainty reduction in these learned maps. Building on these observations, we develop an uncertainty-gated lexicographic algorithm that switches between information gathering and reward exploitation based on perceptual uncertainty. In nectary search, uncertainty gating improves learning relative to exploit-only, $\epsilon$-greedy, fixed-time, and uncertainty-calibrated non-lexicographic controls, with better generalization and robustness to sensory and curvature shifts. These results link information landscapes learned through generative modeling to biological sensing and embodied exploration.
\end{abstract}

\section{Introduction}
A hallmark of biological intelligence is the autonomous seeking of information. Animals move their sensors, reposition their bodies, and select actions that determine what information will become available next, integrating information across visual, olfactory, tactile, and auditory modalities. Efficient active sensing therefore appears to depend on internal representations of how information is distributed across possible actions, yet how such action-conditioned information maps are learned from multisensory interaction with the environment remains poorly understood. Classical work on active sensing has formalized perception--action loops in terms of ideal observers and planners, and recent neuroscience has highlighted the importance of integrating multiple sensory streams during natural behavior. However, much of this literature either studies active sensing in abstract or predominantly unimodal settings, or focuses on multisensory coding without explaining how animals acquire modality-specific maps of where informative actions lie.

Hawkmoth flower foraging provides a particularly rich biological setting for studying this problem. During nectar foraging, moths combine vision and proboscis-derived mechanosensation to inspect flowers, localize the nectary, and track flower motion under changing environmental conditions. Experimental studies have shown that probing behavior depends on corolla geometry \citep{goyret2006mechanosensory, campos2015shape, deora2021learning, deora2021tactile}, that visual flower patterns shape probing trajectories \cite{goyret2010look, kannegieser2024visual}, and that flower-tracking strategies differ across species and light levels \cite{stockl2017comparative, chen2020}. Together, these findings suggest that moths exploit structured, modality-specific distributions of information in the environment, but they do not explain how such information structure is learned or how it is used to guide behavior. \rev{Controlled manipulations of geometry, pattern, and illumination can be related to measured hawkmoth probing and tracking trajectories, providing complementary tests of the same sensing hypothesis.}

In this work, we propose that multimodal generative perception models can provide exactly such a representation. We introduce a predictive-coding framework that learns action-conditioned uncertainty-reduction maps from visual and mechanosensory observations in two hawkmoth tasks: nectary search and flower tracking. The learned maps recover modality-specific information landscapes over action space, reflecting how visual patterns, flower geometry, and contact-dependent mechanosensory cues shape the value of exploratory actions. \rev{When applied to biological trajectories, these maps provide a computational account consistent with nonzero tracking offsets, species- and illumination-dependent tracking, probing along visual guides, and geometry-dependent exploration.} Building on these perception models, we then develop an uncertainty-gated lexicographic control algorithm that switches between information gathering and reward exploitation based on perceptual uncertainty, \rev{improving learning efficiency over fixed-time switching and other controls, including a non-lexicographic policy \finalrev{calibrated for posterior uncertainty}, with higher robustness to sensory degradation under different flower conditions.}

Our contributions are threefold. First, we formulate hawkmoth flower interaction as a multimodal active-sensing problem in which visual and mechanosensory observations are integrated by a variational generative model. Second, we derive spatially interpretable, action-conditioned uncertainty-reduction maps that predict where sensing should occur and explain behavioral structure in animal trajectories. Third, we introduce an uncertainty-gated policy that switches between exploration and exploitation based on the model's perceptual uncertainty. Together, these results connect generative active perception, multisensory integration, and biological behavior in a single embodied framework.

\section{Related work}
\paragraph{Information-guided exploration and active sensing in machine learning.}
A growing body of work uses self-supervised world models, latent uncertainty, and optimism to guide exploration, including methods that seek expected novelty, model disagreement, or task-relevant information about optimal trajectories and useful action abstractions \rev{\citep{sekar2020plan2explore, pathak2019disagreement, sharafeldin2024active, seyde2022love, mehta2022tip, park2023pma}}. Related active-observation and adaptive-sensing methods treat sensing itself as a decision variable, asking when observations should be acquired and how sensing should be coupled to control \citep{nam2021acno, holt2023active, shang2023activevision, treven2024tacos}. \rev{Information-seeking methods include model disagreement \citep{shyam2019active}, latent Bayesian surprise \citep{mazzaglia2022surprise}, prediction-error curiosity \citep{pathak2017curiosity}, and VIME, which augments reward with information gain about dynamics \citep{houthooft2016vime}. Our non-lexicographic control combines task reward with the same perceptual UR signal used by the uncertainty-gated controller. DIAYN \citep{eysenbach2019diayn} instead learns distinguishable skills through skill--state mutual information without task reward; its state-based skill-discovery objective differs from reducing uncertainty over hidden task state.}

\paragraph{Multisensory active perception in robotics.}
Robotics work has begun to exploit cross-modal structure for exploration and manipulation, using tactile prediction errors as curiosity rewards, entropy over physical scene properties to choose where to probe, or stage-dependent sensory reweighting during manipulation \citep{rajeswar2022haptics, haughton2022real, feng2024msbot}. Our work is closest in spirit to these approaches, but differs in three ways: we learn a probabilistic generative perception model rather than only a task policy or scene map; derive modality-specific, action-conditioned \rev{UR maps} over embodied sensing actions; and apply these landscapes to explain animal behavior.

\paragraph{Active sensing, multisensory integration, and predictive coding in neuroscience.}
Neuroscience has long emphasized that perception is active, with animals moving sensory organs and bodies to acquire task-relevant information \citep{kleinfeld2006active, schroeder2010dynamics}. In parallel, multisensory integration studies show that behavior and neural responses combine partially redundant or complementary cues, including visual--haptic integration and context-dependent multisensory enhancement \citep{ernst2002humans, stein2008multisensory, van2014multisensory}. However, these studies often treat active sensing, cue integration, and generative perception separately. Predictive-coding and active-inference theories provide a normative account in which perception and action are organized around generative models \citep{rao1999predictive, friston2010free}; here we instantiate these ideas in multimodal generative models of naturalistic, embodied sensing tasks.

\paragraph{Hawkmoth flower interaction as a model system.}
Hawkmoths provide a natural setting for studying embodied multisensory active sensing. During flower tracking, moths combine visual information from the moving flower with proboscis-derived mechanosensory feedback, with sensory-conflict experiments supporting parallel visual and mechanosensory pathways \citep{roth2016integration}. During nectary search, moths sweep the proboscis across artificial flowers, learn shape-dependent probing strategies, extract tactile features such as curvature and edges, and use visual feedback to refine proboscis placement \citep{deora2021tactile, kannegieser2024visual}. These studies establish the behavioral phenomena we seek to explain, but do not provide a unified computational account of how visual and mechanosensory observations are used to build action-conditioned \rev{UR maps}.

\section{Methods}
\subsection{Nectary search task}

We modeled nectary search as an allocentric, embodied active-sensing task inspired by hawkmoth flower-inspection experiments (Fig.~\ref{fig:fig1}). The environment contains a single artificial flower placed at a random location on a two-dimensional visual canvas. The agent controls a simplified moth head and proboscis tip in canvas coordinates and must locate a small nectary opening at the flower center. This abstraction preserves the key structure of the behavioral setting: moths search over artificial flower surfaces, combine visual and proboscis-derived mechanosensory cues, and are constrained by finite proboscis reach and contact mechanics.

Flower geometry follows the artificial corolla manipulations used in tactile active-sensing studies \citep{deora2021tactile}. Each flower is represented as an axisymmetric surface with central hole radius $r_0$, outer radius $R$, length scale $L$, and curvature parameter $c$, producing geometries from funnel-like to flat surfaces (Fig.~\ref{fig:fig1}a). Flowers are rendered as grayscale top-down images with configurable surface patterns, including unpatterned flowers, crosses, stripes, and circular guides, chosen to approximate visual conditions used in hawkmoth probing studies \citep{goyret2010look, kannegieser2024visual}.

At each time step, the action specifies allocentric target positions for the head and proboscis tip, together with a probing depth. The head is constrained to remain within the canvas, and the proboscis tip is constrained to lie within a fixed-radius disk around the head, enforcing finite reach. The observation $x_t=(x_t^{\mathrm{mech}},x_t^{\mathrm{vis}})$ consists of a head-centered visual crop and a mechanosensory vector encoding contact, normal force, local surface normal, and tangential slip. An episode terminates successfully when the proboscis tip enters the nectary opening with sufficient insertion depth. Full task equations and implementation details are provided in Appendix~\ref{app:task}.

\subsection{Flower-tracking task}

We modeled flower tracking using behavioral trajectories from the hawkmoth tracking experiments of Stöckl et al.~\citep{stockl2017comparative}, in which moths tracked moving artificial flowers under different light conditions. The dataset provides flower and moth position time series for the diurnal \textit{Macroglossum stellatarum} and nocturnal \textit{Deilephila elpenor}. We use these trajectories to define a low-dimensional sensorimotor task in which the latent state $s_t$ is the flower position and the current moth position is treated as the previously executed action $a_{t-1}$.

\begin{figure}
    \centering
    \includegraphics[width=0.9\linewidth]{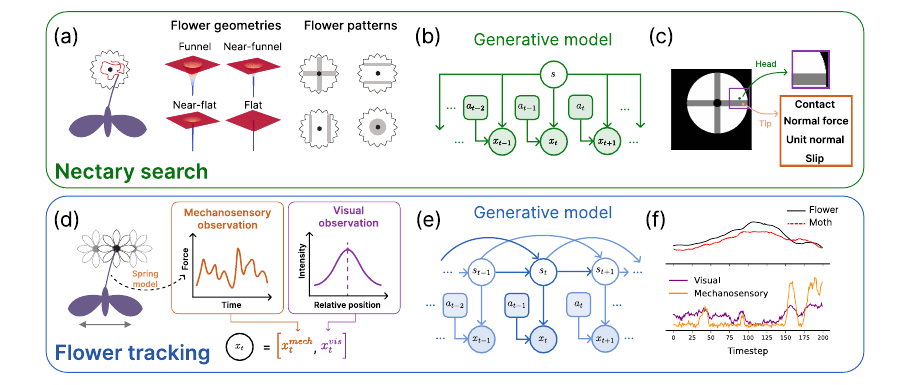}
    \caption{\finalrev{Task setup for nectary search (top) and flower tracking (bottom). (a,d) Task schematics. (b,e) Generative models. (c,f) Sensory inputs.}}
    \label{fig:fig1}
\end{figure}

At each time step, visual and mechanosensory observations are generated from the moth--flower distance (Fig.~\ref{fig:fig1}d). Visual input follows a Gaussian spatial sensitivity profile centered on the flower, whereas mechanosensory input is generated when the flower lies within the reach of a simplified proboscis, producing a noisy thresholded contact-force signal. The empirical trajectories define the initial-state and transition statistics, allowing the task to preserve the temporal structure of real flower tracking while isolating the perceptual problem of integrating visual and proboscis-derived mechanosensory cues. Full observation equations and spring-contact dynamics are given in Appendix~\ref{app:task}.

\subsection{Perception model}
\begin{figure}
    \centering
    \includegraphics[width=0.85\textwidth]{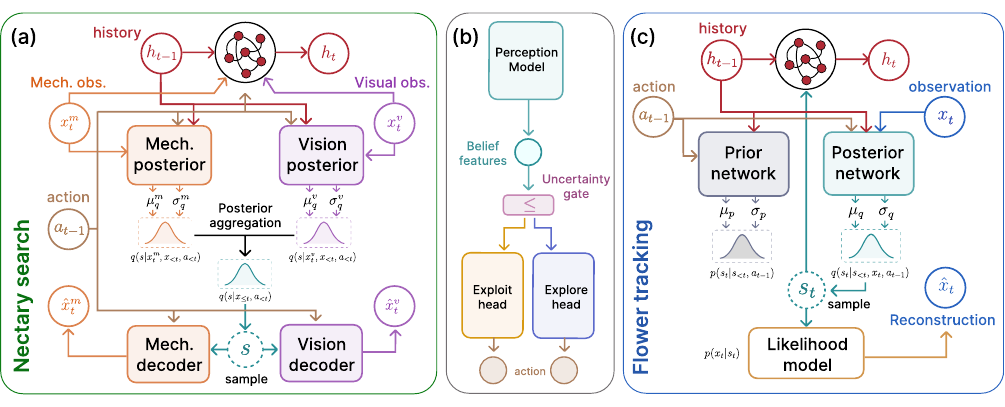}
    \caption{\finalrev{Model architectures. (a) Nectary-search perception. (b) Uncertainty-gated mode switching. (c) Flower-tracking perception.}}
    \label{fig:fig2}
\end{figure}

\paragraph{Nectary search} The perception models for nectary search and flower tracking are based on generative models designed to capture the structure of each task and learned using amortized variational inference \citep{kingma2013auto}. For the nectary search task (Fig. \ref{fig:fig1}b), observations $x_t^\mathrm{vis}$ and $x_t^\mathrm{mech}$ are generated from a stationary latent state $s$, \finalrev{representing hidden environmental properties such as flower position, geometry, and visual pattern. The observed action $a_{t-1}$ specifies the current head and proboscis-tip configuration.} For perception, we assume the trajectory of actions is provided and only the latent state $s$ is being inferred from observations. We also assume that the visual and mechanosensory modalities are conditionally independent given the latent state and action. Then, the action-conditioned joint likelihood can be expressed as
\begin{equation}
    p(x_{1:T}^\mathrm{vis}, x_{1:T}^{\finalrev{\mathrm{mech}}}, s|a_{0:T-1}) = p(s)\prod_{t=1}^{T}p(x_t^\mathrm{vis}|a_{t-1}, s)p(x_t^\mathrm{mech}|a_{t-1}, s).
\end{equation}
The goal of learning a generative model is to maximize the marginal likelihood of observations $p(x_{1:T}|a_{0:T-1})$; however, that requires computation of the true joint posterior, $p(s|x_{1:T}, a_{0:T-1})$, which is intractable. Thus, we approximate the true posterior with an \emph{approximate} variational posterior $q(s|x_{1:T}, a_{0:T-1})$. We further assume that this joint posterior \rev{can be approximated by a mixture of two unimodal posteriors}, $q_\mathrm{mech}$ and $q_\mathrm{vis}$, weighted by learnable weights, $w_\mathrm{mech}$ and $w_\mathrm{vis}$. For example, at some time step $\tau$, the variational posterior can be expressed as
\begin{equation}\label{eqn:posterior_mix}
    q(s|x_{1:\tau}, a_{0:\tau-1}) = w_\mathrm{mech}(\tau)q_\mathrm{mech}(s|x_{1:\tau}^\mathrm{mech}, a_{0:\tau-1}) + w_\mathrm{vis}(\tau)q_\mathrm{vis}(s|x_{1:\tau}^\mathrm{vis}, a_{0:\tau-1}).
\end{equation}
The weights, $w_\mathrm{mech}$ and $w_\mathrm{vis}$, are computed as a function of previous observations and actions. We use a recurrent neural network to summarize the history of observations and actions into a hidden state $h_\tau$, as shown in Figure \ref{fig:fig2}a. The weights are then computed as the \finalrev{softmax} outputs of a neural network, optimized during training, that receives $h_\tau$ as input. Besides its computational benefits, \rev{this modality-specific posterior representation} is also useful for analyzing multisensory integration during active sensing and studying the modality-specific distribution of information in the environment.
In practice, we further approximate the mixture distribution in Eq. \ref{eqn:posterior_mix} by moment matching, yielding a single Gaussian aggregate posterior that can be used with the standard reparameterization and a closed-form Gaussian KL term. Let $\mu_\mathrm{mech}^{\tau}$ and $\sigma_\mathrm{mech}^{\tau}$ be the parameters of $q_\mathrm{mech}$, and $\mu_\mathrm{vis}^{\tau}$ and $\sigma_\mathrm{vis}^{\tau}$ be those of $q_\mathrm{vis}$. We approximate their weighted mixture with the Gaussian
\[
\bar{q}(s|x_{1:\tau},a_{0:\tau-1})
=
\mathcal{N}(\bar{\mu}^{\tau}, \mathrm{diag}((\bar{\sigma}^{\tau})^2)),
\]
where
\begin{equation}
    \bar{\mu}^{\tau}
    =
    w_\mathrm{mech}(\tau)\mu_\mathrm{mech}^{\tau}
    +
    w_\mathrm{vis}(\tau)\mu_\mathrm{vis}^{\tau},
\end{equation}
and the variance is obtained by matching the second moment,
\begin{equation}
    (\bar{\sigma}^{\tau})^2
    =
    w_\mathrm{mech}(\tau)
    \left[
    (\sigma_\mathrm{mech}^{\tau})^2
    +
    (\mu_\mathrm{mech}^{\tau})^2
    \right]
    +
    w_\mathrm{vis}(\tau)
    \left[
    (\sigma_\mathrm{vis}^{\tau})^2
    +
    (\mu_\mathrm{vis}^{\tau})^2
    \right]
    -
    (\bar{\mu}^{\tau})^2 .
\end{equation}
All operations are applied elementwise across latent dimensions since we assume a diagonal covariance for $s$. This gives a single aggregate posterior whose KL divergence to a standard normal prior is available in closed form. The architecture for this perception model is shown in Figure \ref{fig:fig2}a and is trained by maximizing the following $\beta$-weighted , assuming a history-conditioned generative model at every timestep:\begin{equation}\label{eq:elbo}
\begin{aligned}
\mathcal{L}_{\mathrm{NS}}(\theta,\phi)
=
\sum_{t=1}^{T}\rev{\Big\{}
\mathbb{E}_{s\sim\bar{q}}
\big[
&\log p_{\theta}(x_t^\mathrm{vis}|s,a_{t-1})
+
\log p_{\theta}(x_t^\mathrm{mech}|s,a_{t-1})
\big] \\
&-
\beta
D_{\mathrm{KL}}
\left(
\bar{q}_{\phi}(s|x_{1:t},a_{0:t-1})
\,\|\, p(s)
\right)\rev{\Big\}}.
\end{aligned}
\end{equation}

\rev{The aggregate variance reflects both uncertainty within each modality-specific posterior and disagreement between their means. Learned weights can downweight unreliable sensory channels, while $\beta$ balances posterior regularization against reconstruction of both modalities. Training across diverse geometries, visual patterns, contact states, and trajectories further encourages the use of complementary sensory information. In our experiments, the weights vary with sensing context: mechanosensory weight increases during contact, whereas visual weight is larger outside contact (Figure~\ref{fig:fig3}d).}

\paragraph{Flower tracking} In contrast to the nectary search model, the latent state in flower tracking, which corresponds to the flower position, evolves in time. Therefore, the generative model, shown in Figure \ref{fig:fig1}e, has to learn the latent transition dynamics. At each time step, observations are generated based on the distance between the moth position, $a_{t-1}$, and the latent flower position. We write the action-conditioned generative model as
\begin{equation}
    p(x_{1:T},s_{1:T}|a_{0:T-1})
    =
    p(s_1)
    \prod_{t=1}^{T}
    p_\theta(x_t|s_t,a_{t-1})
    \prod_{t=2}^{T}
    p_\theta(s_t|s_{<t}).
\end{equation}
We use an RNN to summarize the history of previous inferred latent states into a hidden state $h_{t-1} \approx s_{<t}$. Because the flower-tracking task is low-dimensional, with one-dimensional actions and observations, we model the full variational joint posterior directly \rev{rather than using modality-specific posteriors as in nectary search}. The transition prior, $p(s_t|s_{<t})$, and variational posterior, $q_\phi(s_t|s_{<t}, x_t, a_{t-1})$, are parameterized as Gaussian distributions whose parameters are output by trained neural networks. The architecture for this model is shown in Figure \ref{fig:fig2}c. The model is trained by maximizing the recurrent ELBO,
\begin{equation}
\begin{aligned}
\mathcal{L}_{\mathrm{FT}}(\theta,\phi)
&=
\sum_{t=1}^{T}\rev{\Big\{}
\mathbb{E}_{s_t\sim q_\phi}
\big[\log p_\theta(x_t|s_t,a_{t-1})
\big] \\
&\quad -
\beta
D_{\mathrm{KL}}
\left(
q_\phi(s_t|s_{<t}, x_t,a_{t-1})
\,\|\, 
p_\theta(s_t|s_{<t})
\right)\rev{\Big\}}.
\end{aligned}
\end{equation}

\subsection{Computing uncertainty reduction maps}\label{sec:ur_maps}
\rev{We compute uncertainty-reduction (UR) maps from the trained perception models to estimate how candidate actions change posterior uncertainty. Each map is conditioned on the observation and action history.} In the nectary search model, the UR map value for an action $a_t'$ is given by
\begin{equation}\label{eqn:ns_ur}
    \Delta H_{\mathrm{NS}}(a_t') = H\big(\bar{q}_\phi(s|x_{1:t}, a_{0:t-1})\big) - H\big(\bar{q}_\phi(s|x_{t+1}', a_t', x_{1:t}, a_{0:t-1})\big),
\end{equation}
where $H(\cdot)$ denotes \rev{differential entropy} and \rev{$x_{t+1}'$} is the counterfactual observation that would be observed if action $a_t'$ was taken. The advantage of the \rev{modality-specific representation} in Eqn. \ref{eqn:posterior_mix} is that we can compute this map separately for the mechanosensory and the visual uncertainty by replacing $\bar{q}_\phi$ with either $q_\mathrm{mech}$ or $q_\mathrm{vis}$, allowing us to analyze the modality-specific information structure of the environment.

For the flower tracking model, the UR map is calculated as
\begin{revision}
\begin{equation}\label{eqn:ft_ur}
    \Delta H_{\mathrm{FT}}(a_t') = H\big(p_\theta(s_{t+1}|s_{\leq t})\big)
    - H\big(q_\phi(s_{t+1}|x_{t+1}',a_t',s_{\leq t})\big).
\end{equation}
For the same future state, the first term is the predictive temporal prior and the second is the posterior conditioned on the candidate observation.
\end{revision}

\subsection{Mode switching algorithm}

\finalrev{Motivated by the UR analysis of biological trajectories, we develop an algorithm} for balancing exploration and exploitation using a lexicographic, mode-switching approach and apply it to the nectary search task. In this algorithm, the belief features from the perception model are used as input for two separate modules: an exploration head and an exploitation head. An uncertainty-based gate decides which module to use for action selection at every time step. This gating is based on a thresholding mechanism such that if the perception model's posterior uncertainty falls below a certain threshold, the exploitation module is used, whereas the exploration module is used whenever uncertainty is above a given threshold. In practice, we implement this gate with hysteresis and use EMA-smoothed posterior uncertainty. More precisely, at time step $t$, the EMA-smoothed uncertainty is computed as
\begin{equation}
    H_\mathrm{ema}(s_t) = \rev{(1-\lambda_\mathrm{ema})}H(s_t) + \rev{\lambda_{\mathrm{ema}}}H_\mathrm{ema}(s_{t-1})
\end{equation}
The exploration module is chosen to act until $H_\mathrm{ema}(s_t)$ falls below a given lower threshold $V_l$, at which point the exploitation module is chosen to act. If uncertainty increases again above an upper threshold $V_h$, the exploration module is chosen again for action. Both action modules are actor-critic models and their output actions are constrained to be within a given delta from the previous action to prevent unrealistic movement. The architecture of this mode-switching model is shown in Figure \ref{fig:fig2}b.

\subsection{Training}\label{sec:training}
\begin{revision}
We apply our mode switching algorithm to the nectary-search task. Perception is first pretrained using smooth trajectories simulating random transitions between a small set of behavior modes, e.g. surface scanning, off-flower cruising, and center probing. This is meant to simulate the smoothness and temporal coherence of real moth trajectories. Examples of such training trajectories are shown in Appendix~\ref{app:example_traj}. For the mode switching model, both the exploration and exploitation networks are trained with Proximal Policy Optimization (PPO) using a clipped surrogate objective \citep{ppo_algorithms} and Generalized Advantage Estimation (GAE) \citep{gae_paper}. \finalrev{With perception frozen, we first pretrain exploration using intrinsic reward $\Delta H_{\mathrm{NS}}(a_t)$ from the observation $x_{t+1}$ following the executed action $a_t$ (Eq.~\ref{eqn:ns_ur}). We then train exploitation on sparse task reward while continuing to train exploration with a lower learning rate; perception remains frozen.} More training and implementation details are included in Appendix~\ref{app:model}.

Our main controller uses \emph{aggregate entropy}: it optimizes the realized reduction in Eq.~\ref{eqn:ns_ur} and gates on $H(\bar q_\phi)$, which includes cross-modal disagreement. We also evaluate a \emph{weighted-unimodal} variant that uses $r_t^{\mathrm{explore}}=\sum_m w_m(t)\Delta H_m(a_t)$, with $m\in\{\mathrm{mech},\mathrm{vis}\}$, and gates on the weighted sum of modality entropies. Thus exploration rewards use reductions in uncertainty, whereas the gate uses the current level of uncertainty. Figures~\ref{fig:fig5} and~\ref{fig:weighted_unimodal} report aggregate-entropy and weighted-unimodal control results, respectively.
\end{revision}

\section{Results}
\subsection{\rev{Multimodal perception models learn structured UR maps in nectary search}}

\begin{figure}
    \centering
    \includegraphics[width=\linewidth]{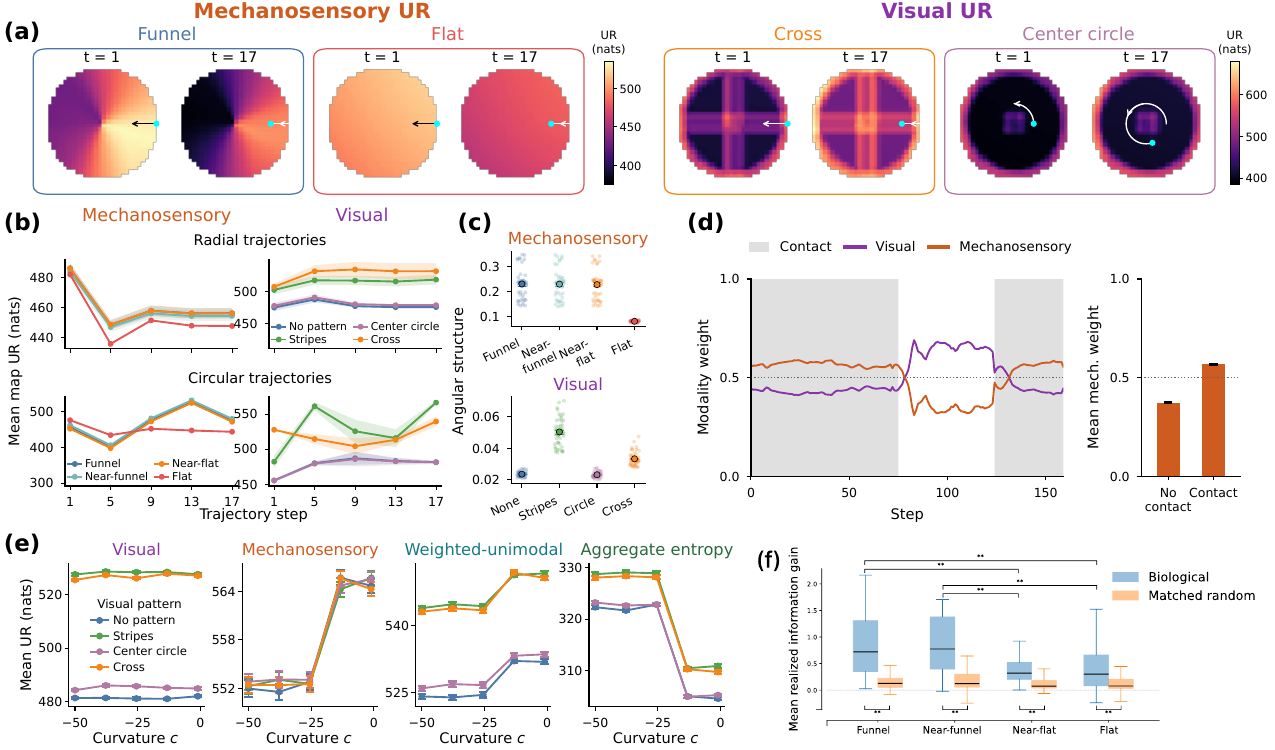}
    \caption{\rev{Uncertainty-reduction maps in nectary search. (a) Mechanosensory and visual UR maps across trajectories and flower conditions. (b) Mean map UR along radial and circular trajectories. (c) Angular anisotropy of final UR maps: variance of mean UR across 16 equal angular sectors around the flower center, weighted by pixel counts and normalized by total map variance. (d) Modality weights along an example trajectory and mean mechanosensory weight with and without contact. (e) Mean visual, mechanosensory, weighted-unimodal, and aggregate-entropy UR across flower curvatures. (f) Realized information gain along biological and matched random trajectories. Shading and error bars in (b--e) show SEM. In (f), $^{**}p<0.01$ (t-tests).}}
    \label{fig:fig3}
\end{figure}

To test whether the learned perception model recovers meaningful action-conditioned information structure, we computed uncertainty-reduction (UR) maps over the flower surface \finalrev{using the modality-specific, weighted-unimodal, and aggregate posterior representations (Sections~\ref{sec:ur_maps} and~\ref{sec:training}).} The resulting maps reveal clear modality-dependent structure that tracks the statistics of the environment. In Figure \ref{fig:fig3}a, mechanosensory UR maps are shown for funnel and flat flowers while the agent follows a radial trajectory. \finalrev{Funnel maps show pronounced spatial structure that changes as the proboscis moves inward, whereas flat-flower maps are more uniform. These patterns reflect the interaction between surface geometry and the trajectory-conditioned belief.} In the same panel, visual UR maps for flowers with cross and center-circle patterns closely reflect the underlying pattern geometry. This is qualitatively consistent with Kannegieser et al. \cite{kannegieser2024visual}, who found that hawkmoths preferentially probe along visual flower patterns, suggesting that the learned visual UR maps recover behaviorally relevant visual structure. These examples suggest that the learned maps capture not only overall uncertainty but also the modality-specific spatial distribution of information.
This structure is further quantified in Figure \ref{fig:fig3}b,c. \finalrev{Mean mechanosensory map UR drops early during radial probing and then changes relatively little, whereas circular probing produces larger temporal fluctuations on curved flowers. Visual UR also depends on the path and surface pattern: cross and stripe patterns yield higher mean map UR than unpatterned and center-circle flowers in these examples.} \rev{Although the corolla is axisymmetric, trajectory history and the current head--tip configuration break rotational symmetry: equal-radius candidate points can produce different contact cues under finite proboscis reach. The anisotropy reflects the conditioned belief and action geometry.} We quantify map structure using angular anisotropy in Figure \ref{fig:fig3}c, which shows that curved flowers produce more anisotropic mechanosensory UR maps than flat flowers, while visual anisotropy varies systematically across visual patterns. Together, these results confirm that the perception model recovers distinct information landscapes for each modality: mechanosensory uncertainty is primarily shaped by corolla geometry, whereas visual uncertainty is primarily shaped by surface pattern.

The multimodal gating network learns corresponding state-dependent integration strategies. As shown in Figure \ref{fig:fig3}d, the relative weights assigned to visual and mechanosensory posteriors vary with the contact state of the proboscis. During contact, the model increases the weight of the mechanosensory posterior, whereas outside contact it relies more heavily on vision. \finalrev{The contact-conditioned averages summarize this shift across geometries and patterns. The model therefore adapts sensory weighting to the observations available during behavior.}

We next asked whether these modality-specific trends are reflected in aggregate information gain across environments. In Figure \ref{fig:fig3}e, we measured the average UR achieved by an agent following random trajectories under different visual patterns and flower curvatures. Visual UR varies strongly across patterns but is largely insensitive to curvature, whereas mechanosensory UR varies strongly with curvature but little with visual pattern. \finalrev{Weighted-unimodal UR combines these dependencies and is larger toward the funnel end of the curvature sweep. Aggregate-entropy UR shows the opposite curvature trend, with larger reductions toward the flatter end. Including disagreement between modality means therefore changes how the combined uncertainty measure varies with geometry.}

Finally, we evaluated whether these learned information landscapes account for biological behavior. In Figure \ref{fig:fig3}f, we computed realized weighted UR along probing trajectories from the behavioral study of Deora et al. \citep{deora2021tactile} and compared them with matched random trajectories that preserved step count and per-step movement magnitude but randomized heading direction. Biological trajectories achieved substantially higher realized information than matched random controls. Moreover, realized UR was highest for funnel and near-funnel flowers and lower for near-flat and flat flowers. This trend mirrors experimental findings that moths spend less time probing funnel-like flowers and more time probing flatter flowers, consistent with the interpretation that curved corollas provide more informative mechanosensory structure. \rev{Together, Figure~\ref{fig:fig3} links modality-specific UR and adaptive sensory weights to biological search. The qualitative visual map structure and behavioral conclusions hold when perception is pretrained on both bright-flower/dark-pattern and reversed-contrast environments (Appendix~\ref{app:model}).}

\begin{figure}[htbp]
    \centering
    \includegraphics[width=0.65\linewidth]{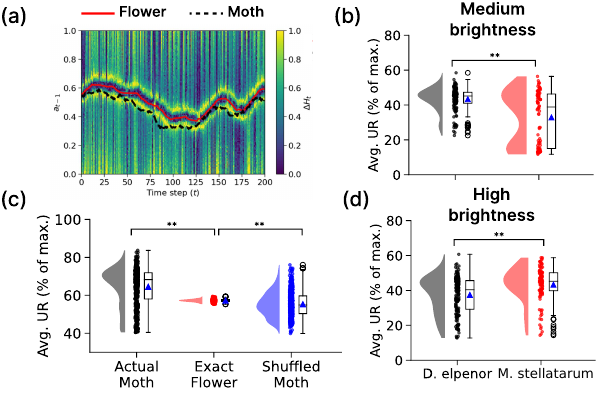}
    \caption{\finalrev{UR in flower tracking. (a) Action-conditioned UR map with flower and moth trajectories. (b,d) Mean realized UR for \textit{D. elpenor} and \textit{M. stellatarum} at medium and high brightness, corresponding to different levels of visual noise. (c) Biological tracking compared with exact flower tracking and shuffled moth trajectories. $^{**}p<0.01$ (t-tests).}}
    \label{fig:fig4}\vspace{-2mm}
\end{figure}

\subsection{\rev{UR maps are consistent with flower-tracking behavior}}

Next, we applied the same predictive-coding framework \finalrev{to} the flower-tracking task by training the perception model directly on biological trajectories from \finalrev{St\"ockl et al.} \citep{stockl2017comparative}. For each trajectory, we used the trained model to compute uncertainty-reduction (UR) maps over the action space as well as the realized UR obtained along the observed moth trajectory. The learned UR maps exhibit a clear bimodal structure around the flower position (Figure \ref{fig:fig4}a), with maximal information gain occurring at small nonzero offsets from the flower rather than at exact alignment. \rev{This structure is consistent with prior analytical predictions \citep{chen2020}. Signal amplitude differs from local informativeness: the visual Gaussian has zero slope at alignment and larger slope at nearby offsets. Saturated mechanosensory force is similarly less sensitive to displacement than force near the contact boundary; $L_{\mathrm{prob}}$ and $d_{\mathrm{sat}}$ shape these peaks. Nonzero offsets can aid sensing and may also reflect sensorimotor delays, phase lag, biomechanical stability, energetic costs, and contact constraints.}

To test whether real moth behavior aligns with these learned information landscapes, we compared the time-averaged realized UR of biological trajectories against two controls: \emph{Shuffled}, where moth and flower trajectories were randomly paired, and \emph{Exact Flower}, where the flower and moth trajectories perfectly match. Real moth
trajectories achieved higher average UR than both
controls (Fig. \ref{fig:fig4}c), suggesting that moths balance precise tracking with information gathering. Finally, comparing species across light conditions showed that the nocturnal \textit{Deilephila elpenor} achieved higher UR at \finalrev{medium brightness} than the
diurnal \textit{Macroglossum stellatarum}\finalrev{,} which performed better under bright light conditions (Fig. \ref{fig:fig4}b, d). \rev{This association is consistent with species-specific sensory ecology.}

\begin{revision}
\subsection{\rev{Uncertainty gating improves learning with condition-dependent generalization}}

We compared five controllers: uncertainty-gated switching, exploit-only, $\epsilon$-greedy random exploration, fixed-time switching, and non-lexicographic control. Fixed-time switching uses the same exploration/exploitation components but switches once at a predetermined step, matched approximately to the adaptive controller's initial exploration duration. The non-lexicographic policy uses $r_t^{\mathrm{task}}+\eta r_t^{\mathrm{explore}}$, with $\eta$ calibrated separately by environment to obtain comparable \finalrev{aggregate-posterior uncertainty}. Settings are given in Appendix~\ref{app:model}.

\finalrev{Figure~\ref{fig:fig5}a shows that adaptive switching learns faster and reaches higher mean success than the baselines in both funnel conditions and on flat flowers with a cross. The cross pattern substantially improves performance on flat flowers, where mechanosensory geometry is weak. Adaptive and non-lexicographic control reach similar posterior entropy levels (bottom row), while their success rates differ in these three conditions. On unpatterned flat flowers, their final success rates are similar.} Fixed-time switching achieves relatively high success on funnel flowers with a cross, but requires longer episodes. Weighted-unimodal control also improves learning relative to its baselines (Figure~\ref{fig:weighted_unimodal}); however, it performs more poorly on flat flowers with a cross, where modality disagreement could inform exploration.

Figure~\ref{fig:fig5}b evaluates controllers trained on cross-pattern funnel and flat flowers. \finalrev{Adaptive switching has the highest mean success under all three shifts for flat flowers. On funnel flowers, adaptive and fixed-time switching achieve similar mean success under curvature shifts.} For funnel flowers, it has a modest advantage under mechanosensory degradation, while other controllers perform better under visual degradation.
\end{revision}

\begin{figure}
    \centering
    \includegraphics[width=\linewidth]{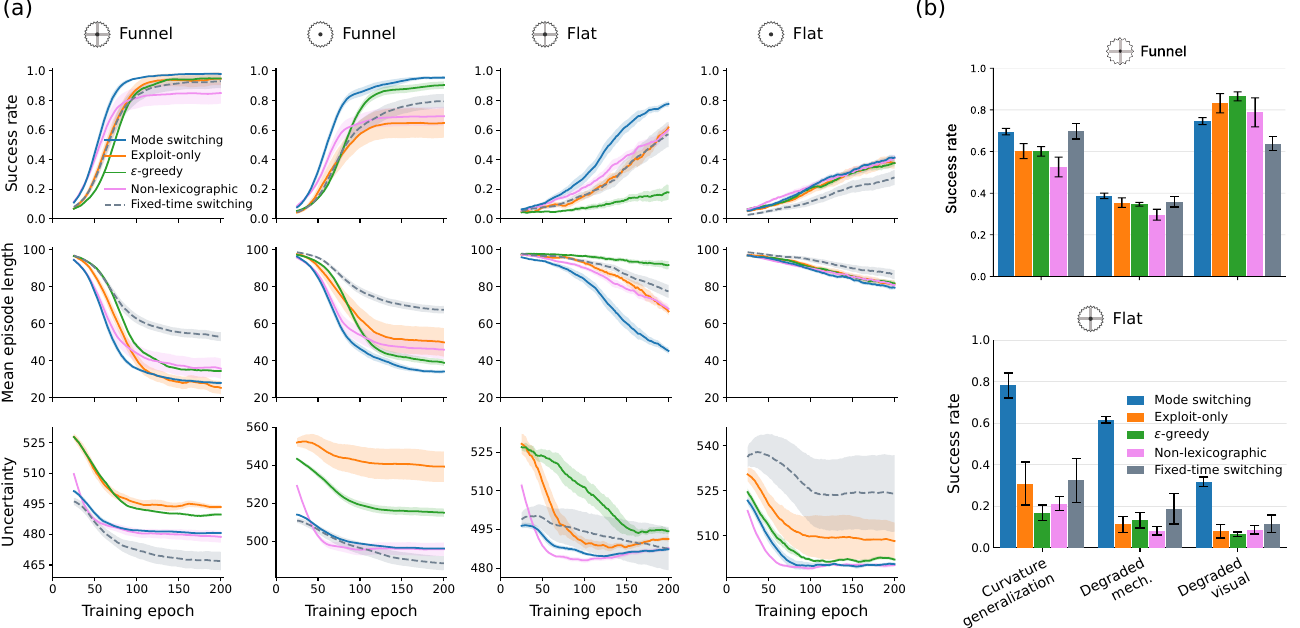}
    \caption{\rev{Aggregate-entropy control and baselines.} (a) Training success rate, mean episode length, and posterior uncertainty across flower curvatures and visual patterns (icons). Curves show 25-epoch moving averages; shading denotes SEM across independent training seeds. (b) Validation success rates (mean $\pm$ SEM) under curvature changes and sensory degradation for cross-pattern flowers.}
    \label{fig:fig5}
    \vspace{-2mm}
\end{figure}

\section{Discussion and Conclusion}
We introduced a multimodal predictive-coding framework for learning action-conditioned information landscapes in embodied active sensing. Across nectary search and flower tracking, the learned uncertainty-reduction (UR) maps recovered structured, modality-specific information: mechanosensory maps reflected corolla geometry and contact state, visual maps reflected surface pattern, and the aggregated posterior combined these sources according to the current sensing regime. When applied to biological trajectories, these maps assigned higher realized information gain to natural behavior than to matched controls and captured qualitative features of hawkmoth sensing, including probing along visual guides, geometry-dependent search, and nonzero tracking offsets. These results suggest that multimodal generative perception models may provide a computational mechanism by which sensory experience gives rise to internal representations of where informative actions lie.

Our uncertainty-gated controller shows how learned UR maps can support downstream behavior through mode switching. Rather than combining intrinsic and extrinsic objectives into a single reward, the policy separates information gathering from reward exploitation and switches between them based on posterior uncertainty. This lexicographic structure is consistent with evidence that animals use distinct exploratory and exploitative movement modes \citep{biswas2023mode}, \rev{and improves learning efficiency over fixed-time switching in all tested training conditions. Comparisons with \finalrev{uncertainty-calibrated} non-lexicographic control and sensory perturbations reveal benefits that depend on geometry and visual context.}

Hawkmoth–flower interactions provide a controlled setting for a broader problem: how embodied agents use actions to gather information from multiple senses. Our framework could be adapted to tasks such as whisker-based object localization or visuotactile manipulation by specifying the relevant latent state, sensory models, and action constraints. The nectary-search and flower-tracking models illustrate how to handle static and changing latent states, respectively. In either setting, reliable uncertainty estimates depend on an appropriate representation of sensory dependencies. \finalrev{Residual sensory dependencies may require a joint observation likelihood, while a joint posterior encoder, as in flower tracking, can represent dependencies during inference.} The controller also uses manually selected uncertainty thresholds; learning these alongside the policies is a natural extension.

Our environments simplify the spatial and sensory demands of natural behavior. Although nectary search uses a five-dimensional action space, spatial coordinates are planar and vision is represented by a head-centered crop. This preserves the dependence of visual information on sensor position while capturing only part of the moth’s visual experience. Extending the environment to full 3D sensing would require richer visual representations and more expensive rendering. Dense uncertainty-reduction maps would also become costly, motivating evaluation over constrained slices or selected candidate actions, as in our nectary-search analysis. Such evaluations require observations for actions beyond those recorded in a behavioral dataset, supplied by an experimental preparation, simulator, or learned forward model.

\subsection*{\finalrev{Code availability}}
\finalrev{Code for simulation, model training, and data analysis: \url{https://github.com/HChoiLab/moth-active-sensing-code}.}

\section*{\finalrev{Acknowledgments and disclosure of funding}}
This work was supported by the Alfred P. Sloan Foundation Fellowship in Neuroscience FG-2022-18969 to HC and the Air Force Office of Scientific Research MURI grant FA9550-22-1-0315 to SS and HC. The funders had no role in study design, data collection and analysis, decision to publish, or preparation of the manuscript. \finalrev{The authors declare no competing financial interests.}

\bibliographystyle{unsrt}
\bibliography{references}

\newpage
\appendix
\setcounter{figure}{0}
\renewcommand{\thefigure}{S\arabic{figure}}
\renewcommand{\theHfigure}{S\arabic{figure}}

\section{Example training and biological trajectories}\label{app:example_traj}
\begin{figure}[h]
    \centering
    \includegraphics[width=\linewidth]{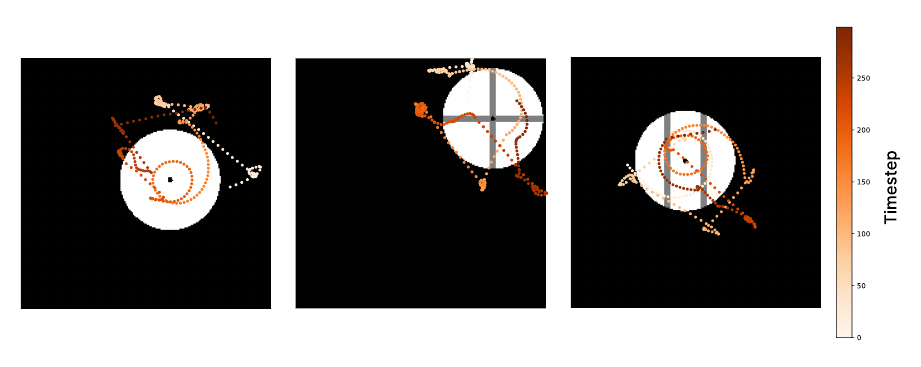}
    \caption{Example training trajectories from the realistic trajectory sampler}
    \label{fig:s1}
\end{figure}

\begin{figure}[h]
    \centering
    \includegraphics[width=\linewidth]{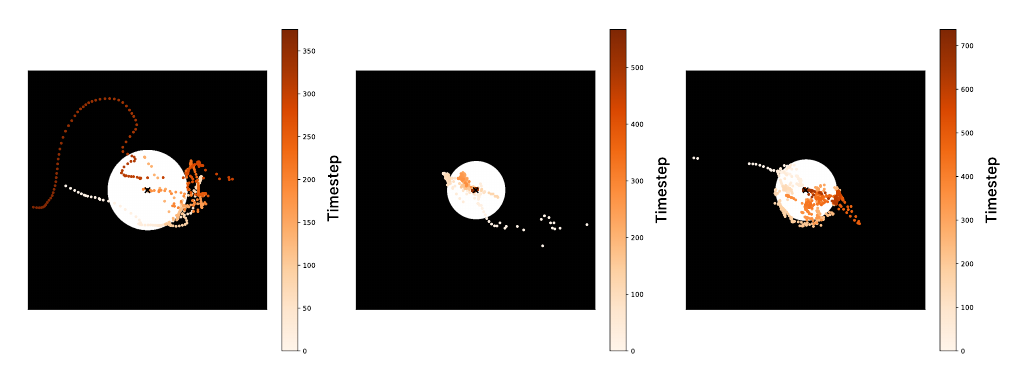}
    \caption{Example behavioral trajectories from the Deora et al. \citep{deora2021tactile} dataset.}
    \label{fig:s2}
\end{figure}

\newpage
\section{\rev{Aggregate-entropy control: additional training metrics}}
\begin{figure}[h]
    \centering
    \includegraphics[width=\textwidth]{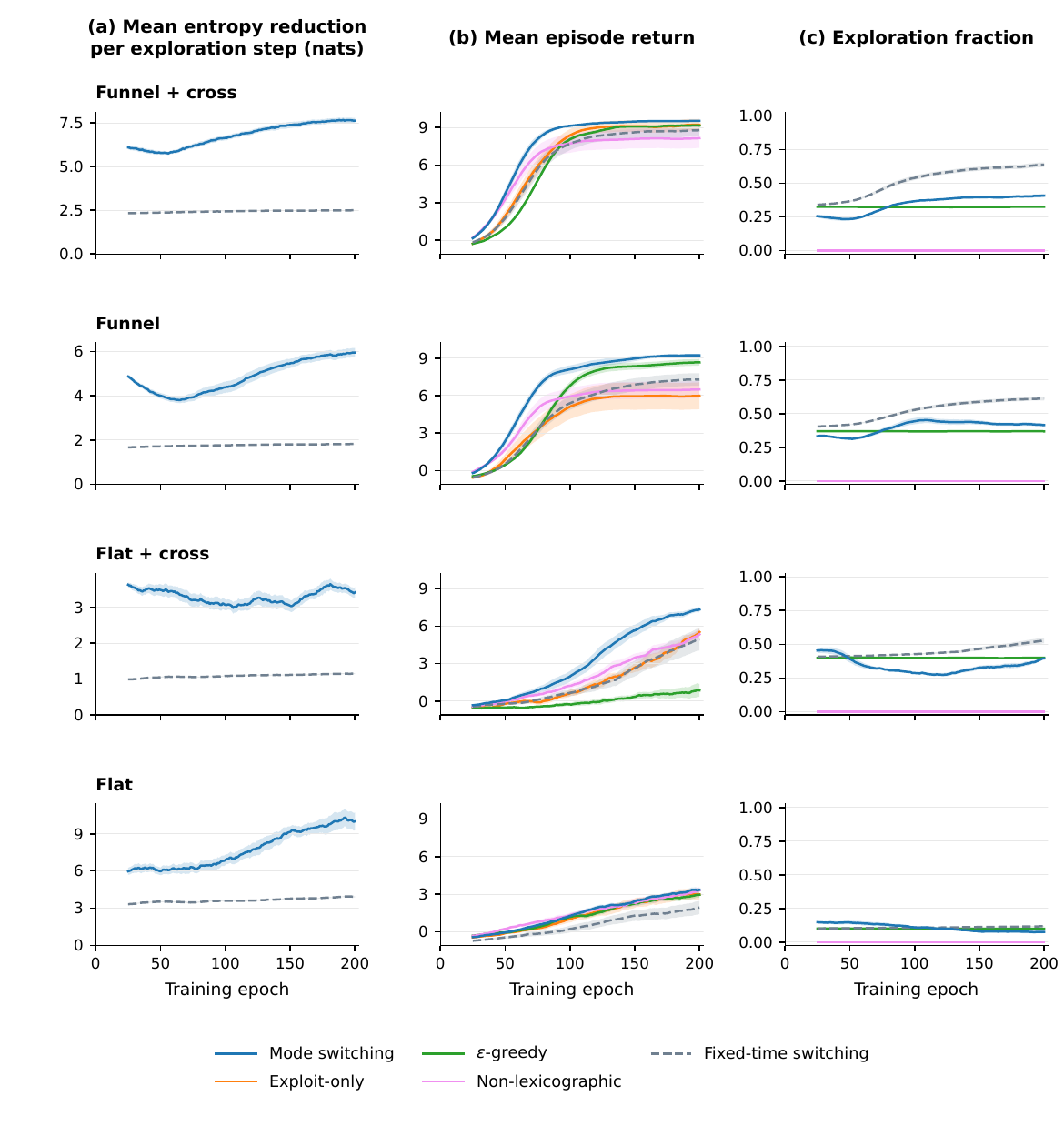}
    \caption{\finalrev{Additional training metrics corresponding to Figure~\ref{fig:fig5}. (a) Mean aggregate-entropy reduction per exploration step for adaptive and fixed-time switching. Only these baselines are plotted here since they use the exploration head. (b) Episode return. (c) Exploration fraction. Shading denotes SEM across independent training runs.}}
    \label{fig:s3}
\end{figure}

\newpage
\section{\rev{Aggregate-entropy control: example mode-switching trajectories}}
\begin{figure}[h]
    \centering
    \includegraphics[width=\textwidth]{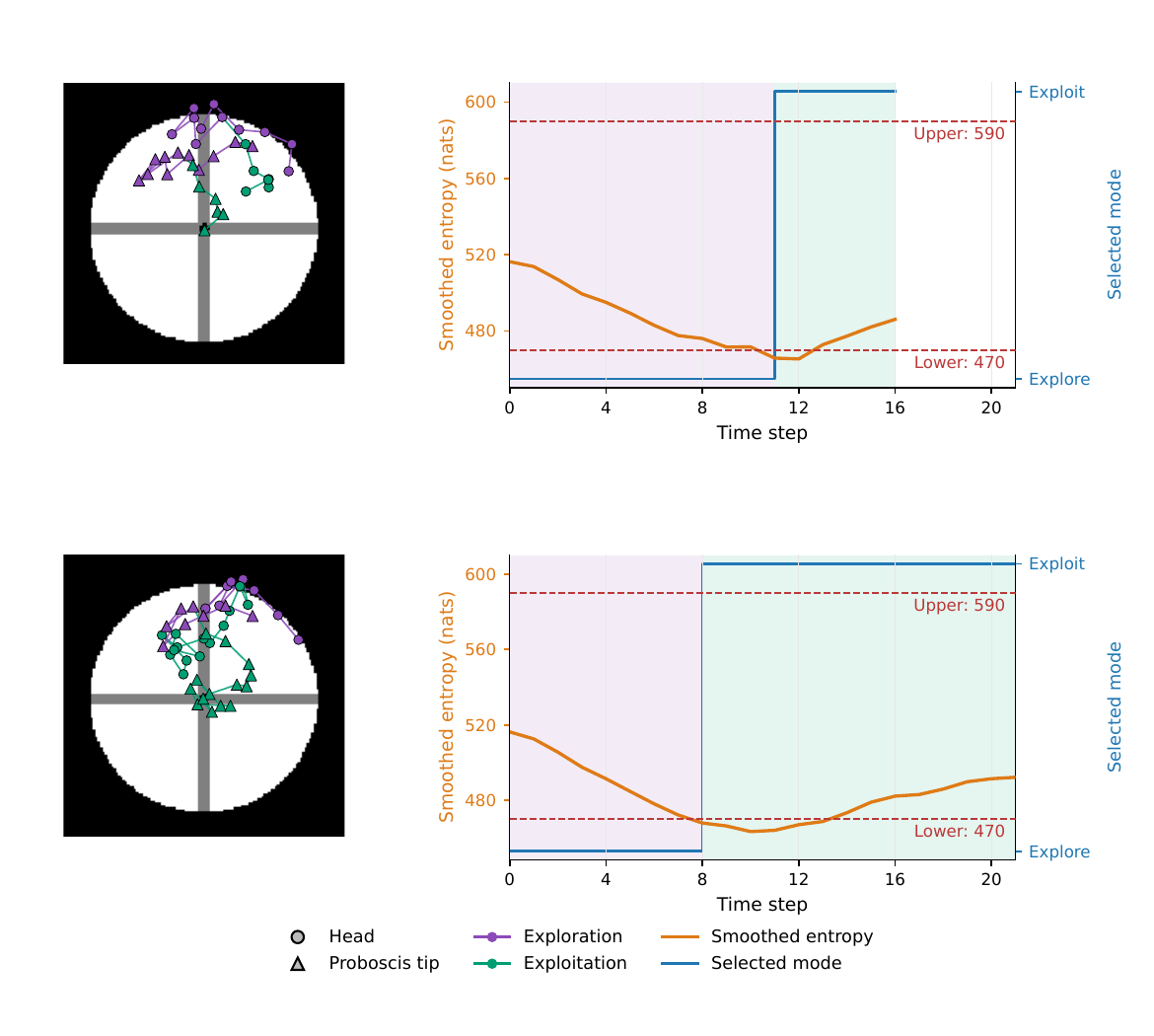}
    \caption{\finalrev{Two successful aggregate-entropy control episodes on a funnel flower with a cross pattern. Left: head (circles) and proboscis-tip (triangles) trajectories, colored purple during exploration and green during exploitation. Right: smoothed aggregate entropy (orange), selected mode (blue), and switching thresholds (dashed), with matching phase shading. The examples switch at steps 11 and 8 and reach the nectary at steps 16 and 21, respectively; steps are indexed from zero.}}
    \label{fig:s4}
\end{figure}

\newpage
\section{Task implementation}\label{app:task}
\begin{finalrevision}
\paragraph{Behavioral data.}
Nectary-search behavioral datasets are from Deora et al.~\citep{deora2021tactile}, available at \url{https://doi.org/10.5061/dryad.sxksn030t}; flower-tracking behavioral datasets are from St\"ockl et al.~\citep{stockl2017comparative}, available at \url{https://doi.org/10.5061/dryad.94kt2}. Both datasets are distributed by Dryad under CC0. We analyze existing datasets and generate sensory observations with the task models below.
\end{finalrevision}
\subsection{Nectary search}
We implemented a Gym-compatible allocentric nectary-search task in which an agent controls a moth head and proboscis tip on a two-dimensional canvas. At the start of each episode, a radially symmetric flower is placed uniformly at a random canvas location such that the full corolla lies inside the image. The flower contains a central nectary hole of radius $r_0$ and outer corolla radius $R$. The proboscis tip is initialized near the outer corolla, with radius sampled from $[0.7R,0.95R]$ and angle sampled uniformly from $[-\pi,\pi]$; the head is initialized outward from the tip at $0.8$ times the proboscis length.

The action at each step is
\[
a_t = (h_x,h_y,p_x,p_y,d) ,
\]
where $(h_x,h_y)$ and $(p_x,p_y)$ are normalized absolute canvas coordinates for the head and proboscis tip, and $d$ is the insertion or pressing depth. Head coordinates are clipped to the canvas. The tip target is denormalized and projected into a disk of radius equal to the proboscis length around the head, enforcing reachability.

Observations are a dictionary containing mechanosensory and visual inputs. The mechanosensory vector is
\[
x_t^{\mathrm{mech}} = (c_t, f_t, n^x_t, n^y_t, s_t),
\]
where $c_t$ indicates contact, $f_t$ is a log-scaled normal-force proxy, $(n^x_t,n^y_t)$ is the local projected surface normal, and $s_t$ is a log-scaled tangential slip proxy. Contact is present only when the tip is inside the corolla, outside the nectary hole, and pressing with positive depth. The corolla surface is modeled as an axisymmetric height field
\[
z(r) = L \left(\frac{r-r_0}{R}\right)^{\exp(c)},
\]
with a flat surface obtained when $c=-\infty$. Surface normals are computed analytically from $\partial z / \partial r$. The normal-force proxy is proportional to clipped penetration depth and normalized by a maximum pressing depth.

The visual observation is a head-centered grayscale crop from a top-down canvas image. The rendered flower has configurable background, flower, hole, and pattern intensities. Patterns include no pattern, a cross, paired stripes, or a central circular marking; the hole is rendered last so that it remains visually distinct. Crops cover a square field of view of size $v \times v$ pixels where $v$ is the visual field size, and are padded with the background value outside the canvas.

An episode terminates successfully when the proboscis tip lies inside the nectary hole and the insertion depth exceeds the required success depth. The reward is sparse:
\[
r_t =
\begin{cases}
10, & \text{if the nectary is reached with sufficient depth},\\
-0.01, & \text{otherwise}.
\end{cases}
\]
Episodes are truncated after a fixed maximum number of steps.

\subsection{Flower tracking}
We constructed a one-dimensional flower-tracking task from digitized moth-flight recordings obtained from \finalrev{St\"ockl et al.} \citep{stockl2017comparative}. Each raw file contained tracking data from 100 Hz videos, with each trial represented by four columns: flower $x/y$ position followed by moth $x/y$ position. We use only $y$-axis coordinates since that is the direction of movement in the experiments:
\[
s_t = y^{\mathrm{flower}}_t, \qquad a_{t-1} = y^{\mathrm{moth}}_t ,
\]
where $s_t$ is treated as the latent flower state and \finalrev{$a_{t-1}$} as the moth/sensor position.

Visual observations are generated synthetically from the relative distance between the flower and moth:
\[
\Delta_t = |s_t-a_{t-1}|,
\]
\[
\mu^{\mathrm{vis}}_t =
A \frac{1}{\sqrt{2\pi\nu^2}}
\exp\left(-\frac{\Delta_t^2}{2\nu^2}\right),
\qquad
x^{\mathrm{vis}}_t =
\mu^{\mathrm{vis}}_t + \epsilon_t,
\quad
\epsilon_t \sim \mathcal{N}(0,\sigma^2_\mathrm{vis}).
\]
Thus the visual signal is strongest when the moth is aligned with the flower and decreases with a Gaussian tuning width $\nu$. 

Mechanosensory observations are simulated as a one-dimensional proboscis contact signal. Let
\[
d_t = |s_t-a_{t-1}| - L_{\mathrm{prob}}
\]
be the signed gap between the flower and the proboscis reach. Contact force is modeled as a saturated spring under compression,
\[
F^{\mathrm{raw}}_t =
k_p \, \mathrm{clip}(-d_t,0,d_{\mathrm{sat}}),
\]
followed by first-order low-pass filtering,
\[
F^p_t = F^p_{t-1} + \frac{\Delta t}{\tau_p}
\left(F^{\mathrm{raw}}_t - F^p_{t-1}\right).
\]
The filtered force is converted into a rectified firing-rate observation,
\[
r_t = \alpha \max(F^p_t - F_{\mathrm{thr}},0),
\]
with Gaussian noise added to yield the final mechanosensory observation:
\begin{finalrevision}
\[
x_t^\mathrm{mech}\sim\mathcal{N}(r_t,\sigma^2_\mathrm{mech}).
\]
\end{finalrevision} The final multisensory observation is the two-channel signal
\[
x_t = \left(x^{\mathrm{vis}}_t, x^{\mathrm{mech}}_t\right).
\]

\newpage
\section{Model implementation and training details}\label{app:model}

\paragraph{Perception model.}
\rev{For aggregate-entropy control (Figure~\ref{fig:fig5}),} we used a recurrent multimodal variational model that receives the mechanosensory vector $x_t^\mathrm{mech}$, the head-centered visual crop $x_t^\mathrm{vis}$, and the executed action \rev{$a_{t-1}$}. The mechanosensory observation is embedded with a linear layer, and the visual crop is embedded with a small convolutional encoder consisting of two convolutional layers with group normalization and SiLU nonlinearities, followed by adaptive average pooling and a linear projection. \rev{The mechanosensory and visual embeddings have dimensions 16 and 32, respectively. Each modality embedding and the action are passed through a separate tanh RNN with hidden size 512.}

The latent state $s$ was modeled as a diagonal Gaussian with latent dimension \rev{512}. We used a mixture-of-experts posterior over the visual and mechanosensory encoders, with learned mixture weights and moment matching to obtain the aggregate Gaussian posterior. The decoder reconstructed both modalities from $(s, a_{t-1})$: mechanosensory contact was decoded with a Bernoulli likelihood for the binary contact channel and Gaussian likelihoods for the continuous channels (e.g. normal force, normal vector components, and slip), while the visual crop was decoded with a Gaussian pixel likelihood with \rev{fixed variance}. The perception model was trained with the $\beta$-VAE objective in Eq.~\ref{eq:elbo} with \rev{$\beta=0.003$}.

\paragraph{Uncertainty-gated policies.}
The uncertainty-gated model augmented the perception model with two actor-critic heads: an exploration policy and an exploitation policy. Both policies operated on belief features
\[
b_t = [\mu_t,\ell_t,h_t],
\]
where $\ell_t$ denotes the posterior log-variance used for uncertainty estimation. Each actor and critic was a two-layer MLP with 128 hidden units per layer and ReLU activations. Continuous actions were represented with squashed Gaussian policies and constrained to the environment action bounds and local action-delta bounds. The exploitation policy was trained from the sparse task reward, while the exploration policy was trained from realized \rev{UR} rewards.

\paragraph{Training.}
\begin{revision}
\finalrev{Perception is pretrained on trajectories combining surface scanning, off-flower cruising, and center probing across geometries and visual patterns. Training includes both bright-flower/dark-pattern and reversed-contrast environments. We then freeze perception, pretrain exploration using realized aggregate-entropy reduction, and train exploitation on sparse task reward while continuing exploration learning with a lower learning rate. Flower-tracking perception is trained separately from biological trajectories and simulated sensory observations, without policy heads.}
\end{revision}

Policy heads were trained online on environment rollouts using Adam. \rev{Policy training uses 200 epochs, a batch size of 16, and a maximum episode length of 100 steps.} Actor-critic updates used PPO with discount $\gamma=0.99$, GAE parameter $\lambda=0.95$, clipping parameter $0.2$, value-loss coefficient $0.5$, and entropy coefficient $10^{-3}$. We trained models using NVIDIA Quadro RTX 6000. \rev{Perception pretraining took about 15 minutes for 300 epochs, while policy training took about 10 minutes for 200 epochs.}

\begin{revision}
\finalrev{The exploitation learning rate is $10^{-4}$. The exploration head uses a learning rate of $10^{-4}$ during pretraining and $7\times10^{-5}$ during switching training. The $\epsilon$-greedy random-action probabilities are $0.325$, $0.37$, $0.4$, and $0.1$ for funnel-cross, funnel, flat-cross, and flat, respectively.} PPO and GAE coefficients are as above. The gate uses $\lambda_{\mathrm{ema}}=0.85$; Table~\ref{tab:aggregate_runs} lists its central threshold and hysteresis half-width, giving $V_l=\text{gate}-\text{hysteresis}$ and $V_h=\text{gate}+\text{hysteresis}$. The table also lists fixed switch times and non-lexicographic reward weights.
\end{revision}

\begin{table}[ht]
\centering
\begin{revision}
\caption{Settings for the control experiments in Figure~\ref{fig:fig5}.}
\label{tab:aggregate_runs}
\small
\begin{tabular}{lrrrr}
\toprule
Environment & Gate & Hysteresis & Fixed step & $\eta$\\
\midrule
Funnel + cross & 530 & 60 & 33 & 0.07\\
Funnel & 535 & 60 & 40 & 0.05\\
Flat + cross & 520 & 50 & 40 & 0.0065\\
Flat & 550 & 52 & 10 & 0.004\\
\bottomrule
\end{tabular}
\end{revision}
\end{table}

\begin{revision}

\paragraph{Map computation and extensions.}
\finalrev{UR Maps use a $30\times30$ candidate grid at trajectory steps 1, 5, 9, 13, and 17, with probing depth fixed at 0.1. Angular anisotropy is computed from each final map and pooled across paths and patterns for mechanosensation, and across paths and geometries for vision. Contact-conditioned weights are averaged within each episode before computing their mean and SEM. Figure~\ref{fig:fig3}(e) uses 200 independently sampled 100-step random trajectories per pattern--curvature condition; its SEM is calculated over stepwise UR values. Map summaries use all valid grid points; panel (a) masks the area outside the flower for display.}

Nectary-search maps evaluate a two-dimensional slice of the five-dimensional action space: probing depth is fixed and head position has a fixed offset from the candidate tip position. Extending the environment to 3D perception could similarly evaluate planes or sampled candidate actions, avoiding exhaustive enumeration. Applying UR maps to a new dataset requires an experiment, simulator, or learned forward model that supplies counterfactual observations beyond recorded trajectories.

Our modeling approach is not limited to two modalities. For $K$ modalities, a $K$-way softmax supplies weights $w_k$, and diagonal moment matching extends as
\[
\bar\mu=\sum_{k=1}^{K}w_k\mu_k,\qquad
\bar\sigma^2=\sum_{k=1}^{K}w_k(\sigma_k^2+\mu_k^2)-\bar\mu^2.
\]
Therefore, aggregation scales linearly with $K$ at fixed latent dimension; however, sensory encoding and candidate-action evaluation incur their own computational costs. \finalrev{Residual cross-modal dependencies can be represented with a joint observation likelihood and a joint posterior encoder.}
\end{revision}

\begin{finalrevision}
\paragraph{Evaluation and statistical summaries.}
Figures~\ref{fig:fig5}a and~\ref{fig:s3} show means and sample SEM across independent training runs after applying a trailing 25-epoch moving average within each run. Validation uses ten episodes per condition. Curvature generalization tests use $c\in\{-2,-5,-20,-50\}$; mechanosensory degradation tests remove surface-normal channels, add Gaussian noise with standard deviation 0.1, or blank the mechanosensory input. Visual degradation tests shift the cross away from the flower center, add Gaussian noise with standard deviation 0.1, or blank vision. Success is averaged across conditions within each test suite and then across trained models. Figure~\ref{fig:s3}a averages entropy reduction only over exploration-selected steps; its other columns average over complete episodes. Box plots in Figures~\ref{fig:fig3}f and~\ref{fig:fig4} show medians and interquartile ranges; double asterisks indicate $p<0.01$ with Student's t-tests.
\end{finalrevision}

\begin{figure}[!ht]
    \centering
    \includegraphics[width=\linewidth]{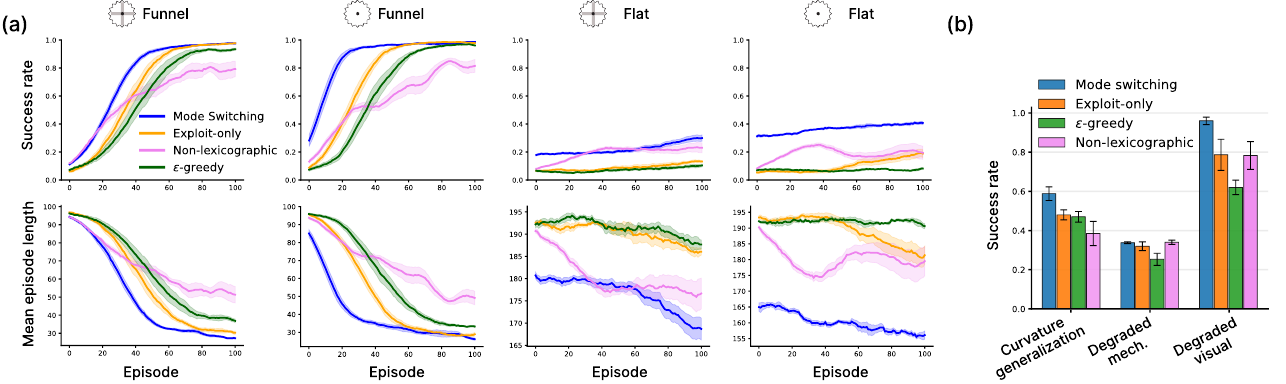}
    \caption{\rev{Weighted-unimodal control. (a) Training success and episode length in four flower conditions. (b) Curvature and sensory-shift validation after funnel-cross training.}}
    \label{fig:weighted_unimodal}
\end{figure}

\end{document}